\documentclass{bmvc2k}

\pdfoutput=1

\usepackage{hyperref}
\usepackage{epsfig}
\usepackage{graphicx}
\usepackage{amsmath}
\usepackage{amssymb}
\usepackage{verbatim}
\usepackage{gensymb}
\usepackage{textcomp}
\usepackage{bm}
\usepackage{makecell}
\usepackage{algorithm}
\usepackage[noend]{algpseudocode}
\usepackage{wrapfig,booktabs}
\usepackage{color}
\usepackage[normalem]{ulem}
\usepackage{pifont}
\usepackage{multirow}
\usepackage{mathtools}
\usepackage{dsfont}
\usepackage{comment}

\newcommand{\rdalgoms}{R2D2$_{s_3}$}

\newcommand{\vsn}{VSM}
\newcommand{\vsnfull}{\textit{View Synthesis Module}}

\newcommand{\vslfull}{\textit{View Synthesis Loss}}
\newcommand{\gtefull}{\textit{Grid Transformation Encoder}}
\newcommand{\dsnfull}{\textit{Depthmap Synthesis Network}}
\newcommand{\gte}{GTE}
\newcommand{\dsn}{DSN}
\newcommand{\dadloss}{cm}
\newcommand{\dadlossfullname}{\textit{Contrastive Matching Loss}}

\title{ViewSynth: Learning Local Features from Depth Using View Synthesis}

\addauthor{Jisan Mahmud$^{\dagger}$}{jisan@cs.unc.edu}{1}
\addauthor{Rajat Vikram Singh}{singh.rajat@siemens.com}{2}
\addauthor{Peri Akiva$^{\dagger}$}{peri.akiva@rutgers.edu}{3}
\addauthor{Spondon Kundu}{spondon.kundu@siemens.com}{2}
\addauthor{Kuan-Chuan Peng$^{\dagger}$}{kpeng@merl.com}{4}
\addauthor{Jan-Michael Frahm}{jmf@cs.unc.edu}{1}

\addinstitution{University of North Carolina at Chapel Hill, Chapel Hill, NC}
\addinstitution{Siemens Corporate Technology, Princeton, NJ}
\addinstitution{Rutgers University, New Brunswick, NJ}
\addinstitution{Mitsubishi Electric Research Laboratories, Cambridge, MA}

\runninghead{}{}

\def\etal{\emph{et al}\bmvaOneDot}

\begin{document}
\maketitle
\let\thefootnote\relax\footnotetext{$^{\dagger}$This work was done when Jisan Mahmud and Peri Akiva were interns and Kuan-Chuan Peng was a staff scientist at Siemens Corporate Technology.}
\begin{abstract}
The rapid development of inexpensive commodity depth sensors has made keypoint detection and matching in the depth image modality an important problem in computer vision.
Despite great improvements in recent RGB local feature learning methods, adapting them directly in the depth modality leads to unsatisfactory performance.
Most of these methods do not explicitly reason beyond the visible pixels in the images.
To address the limitations of these methods, we propose a framework ViewSynth, to jointly learn: (1) viewpoint invariant keypoint-descriptor from depth images using a proposed \textit{Contrastive Matching Loss}, and (2) view synthesis of depth images from different viewpoints using the proposed \vsnfull~and \vslfull.
By learning view synthesis, we explicitly encourage the feature extractor to encode information about not only the visible, but also the occluded parts of the scene.
We demonstrate that in the depth modality, ViewSynth outperforms the state-of-the-art depth and RGB local feature extraction techniques in the 3D keypoint matching and camera localization tasks on the RGB-D dataset 7-Scenes, TUM RGBD and CoRBS in most scenarios.
We also show the generalizability of ViewSynth in 3D keypoint matching across different datasets.
\end{abstract}

\vspace{-4.5mm}
\section{Introduction}
\vspace{-1.2mm}

Accurate local feature correspondence matching is a crucial step in many computer vision applications like structure-from-motion and multi-view stereo \cite{schonberger2016structure,agarwal2011building,heinly2015reconstructing,schonberger2016pixelwise}, image retrieval \cite{noh2017large,babenko2015aggregating,wang2010robust}, geo-localization \cite{mshah2014,jin2015predicting,kim2017learned}, camera localization \cite{sattler2018benchmarking,taira2018inloc,dusmanu2019d2}, and object pose estimation \cite{georgakislearning}.
Most of these applications utilize RGB based local features.
Unlike RGB images, depth images contain 3D information \cite{matusiak2018depth}, and they are invariant to color, texture, and illumination changes \cite{wang2018convolutional,liu2016depth}.
These properties have encouraged recent studies that use depth images in 3D correspondence matching \cite{georgakis2018end,liu2018fast,matusiak2018depth}, object pose estimation\cite{georgakislearning}, human pose estimation \cite{mccoll2011human,wang2016human}, etc.
Depth images are especially suitable for establishing local feature correspondences \cite{georgakis2018end,liu2018fast,matusiak2018depth} when high color, texture, or illumination variation is expected,
which motivates us to propose a framework for learning keypoints and descriptors from depth images towards the tasks of 3D keypoint matching and camera localization.

Most of the handcrafted feature based \cite{sift,rublee2011orb,matusiak2018depth,liu2018fast}, and deep-learning based \cite{yi2016lift,ono2018lfnet} keypoint-descriptor extraction methods take a \textit{detect-then-describe} (\textit{DtD}) approach, where keypoints and descriptors are estimated separately.
D2Net \cite{dusmanu2019d2} shows that a \textit{detect-and-describe} (\textit{DaD}) approach - where detection and description are jointly estimated, and leads to better performance.
However, despite the state-of-the-art (SOTA) performance in the RGB modality, \cite{dusmanu2019d2} is not directly applicable in the challenging depth image modality due to model collapse \cite{wu2017sampling}, which we explain in Sec \ref{kplearningsubsection}.
Moreover, most keypoint-descriptor learning methods are designed to primarily learn from the commonly visible parts of RGB \cite{dusmanu2019d2,revaud2019r2d2,ono2018lfnet} or depth \cite{georgakis2018end,georgakislearning} image pairs.
They do not explicitly enforce encoding information about the occluded parts of the scene.
In contrast, we propose to explicitly enforce the learned features (which are used to generate keypoints and descriptors) to be able to synthesize the depth from different viewpoints, including the areas occluded in the original viewpoint.
Therefore, the learned features are trained to include the information beyond the visible pixels.
Sitzmann \etal \cite{sitzmann2019scene} showed that learning 3D-structure-aware scene representation can improve various tasks like few-shot reconstruction, shape and appearance interpolation, novel view synthesis, etc.
Inspired by \cite{sitzmann2019scene}, we hypothesize that learning view synthesis of depth images will encourage encoding such 3D-structure-aware information, which would be useful for extracting keypoints and descriptors that can be correctly matched.

\begin{figure}[t!]
\centering
\includegraphics[width=1.0\linewidth]{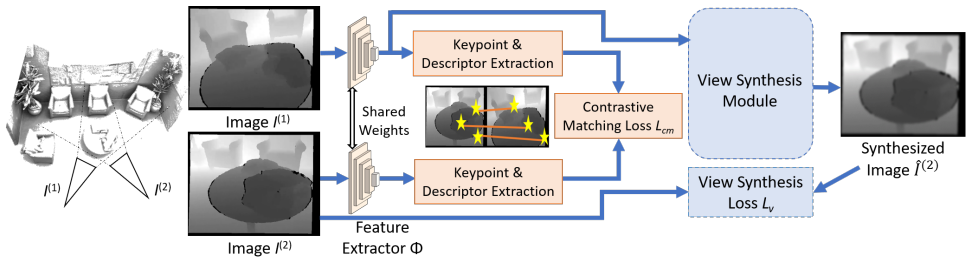}
\vspace{-9mm}
\caption{
The ViewSynth framework.
Dense features, keypoints and descriptors are extracted from depth images $I^{(1)}, I^{(2)}$.
\dadlossfullname~supervises keypoint and descriptor learning.
Simultaneously, \vsnfull~trained with \vslfull~synthesizes depth image from $I^{(2)}$'s view from $I^{(1)}$'s features.
}
\vspace{-5mm}
\label{fig:teaser}
\end{figure}

To this end, we propose a local feature learning framework \textbf{ViewSynth} (Figure \ref{fig:teaser}) which (1) jointly estimates keypoints and descriptors with a \textit{DaD} approach, and (2) explicitly enforces encoding information beyond the visible pixels by learning view synthesis.
First, we propose a \dadlossfullname, $L_{\dadloss}$, which uses a contrastive loss coupled with the hardest negative sampling \cite{felzenszwalb2008discriminatively} to learn viewpoint invariant keypoint-descriptor, while circumventing the model collapse  \cite{wu2017sampling} problem.
We also propose the \vsnfull~(\vsn), which takes in a depth image and a relative pose, and synthesizes the depth image from that relative pose, and \vslfull, $L_v$ to train \vsn.
\vsn~consists of: the \gtefull~(\gte), which encodes the transformation-related parameters between the images, and the \dsnfull~(\dsn), which uses the output of \gte~and dense features of one depth image to synthesize the other depth image.

In summary, we make the following contributions:
(1) we propose the \textbf{ViewSynth} framework, which learns view synthesis using the \vsnfull~(\vsn) (composed of the \gtefull~(\gte) and the \dsnfull~(\dsn)) towards improving keypoint matching,
(2) the \dadlossfullname, $L_{\dadloss}$, for learning keypoints and descriptors in depth images,
(3) the \vslfull, $L_v$, to train \vsn.
ViewSynth outperforms the SOTA depth \cite{georgakis2018end} and SOTA RGB \cite{dusmanu2019d2,revaud2019r2d2} local feature methods in the 3D keypoint matching and camera localization tasks on RGB-D dataset 7-Scenes \cite{msr7_2013}, TUM \cite{sturm12iros}, and CoRBS \cite{wasenmuller2016corbs} datasets, while showing better generalizability across datasets.

\section{Related work}
\textbf{Sparse local feature learning:}
While traditional methods like SIFT \cite{sift}, SURF \cite{surf} and ORB \cite{rublee2011orb} use handcrafted features for extracting keypoint-descriptor, deep-learning based methods \cite{ono2018lfnet,tian2017l2,dusmanu2019d2,revaud2019r2d2} outperform the traditional ones by learning features from the images.
Most of these methods take a \textit{DtD} approach \cite{sift,surf,rublee2011orb,calonder2010brief,yi2016lift,ono2018lfnet,detone2018superpoint}.
In contrast, D2Net \cite{dusmanu2019d2} and R2D2 \cite{revaud2019r2d2} propose \textit{DaD} approaches that share all \cite{dusmanu2019d2}, or most \cite{revaud2019r2d2} parameters between keypoint detection and description, and achieve the SOTA keypoint matching performance in RGB modality.
Our experiments show that these methods are either not trainable, or underperform when directly adapted in the depth modality.
Moreover, none of these methods explicitly seek to encode information beyond the visible pixels in the images, towards keypoint-descriptor extraction.
In contrast, ViewSynth explicitly learns to encode this information by learning view synthesis, and shows that this is beneficial for keypoint matching.
Suwajanakorn \etal \cite{Suwajanakorn2018} proposed 3D keypoint learning from RGB images via geometric reasoning to obtain a fixed number of keypoints for pose estimation.
Their method only detects keypoints without providing their descriptors, but ViewSynth can provide both.

\textbf{Learning from depth data:}
The reliance on depth data has recently seen a surge in applications like 3D object detection \cite{drost20123d,qi2018frustum}, facial emotion recognition \cite{Szwoch2015FacialER}, people counting \cite{iguernaissi2018people,bondi2014real}, activity recognition \cite{jalal2017robust,liu2016depth} and human pose estimation \cite{mccoll2011human,wang2016human}.
Recently, the authors of \cite{georgakislearning} learn modality-invariant keypoints between RGB and rendered depth images for object pose estimation.
Georgakis \etal \cite{georgakis2018end} learn keypoints and descriptors from depth images towards 3D correspondence matching.
Most depth image keypoint-descriptor methods either take a \textit{DtD} approach \cite{matusiak2018depth,zeng20173dmatch,liu2018fast}, or do not share all parameters between detection and description \cite{georgakislearning,georgakis2018end}.
D2Net \cite{dusmanu2019d2} showed that a \textit{DaD} approach that shares all parameters between detection and description outperforms the \textit{DtD} methods, or methods that do not share all parameters.
In addition, these methods also do not explicitly seek to encode information about the occluded parts of the images, which ViewSynth is designed to overcome. %

\textbf{Synthesizing novel views:}
The use of view synthesis has been largely focused on generating information missing in current views with known applications in point cloud reconstruction by depth image synthesis \cite{lin2018learning}, depth image super-resolution \cite{song2018deeply}, layered 3D scene inference \cite{tulsiani2018layer}, image inpainting \cite{yang2017high,pathak2016context} and image-to-image translation \cite{isola2017image,choi2019extreme}.
Zhou \etal \cite{zhou2016view} predicts an appearance flow to synthesize novel views from an image.
Sitzmann \etal \cite{sitzmann2019scene} learns to implicitly represent a scene in a 3D-structure-aware manner, towards novel view synthesis, shape and appearance interpolation, and few-shot reconstruction.
These methods typically do not generate keypoints or descriptors for matching.
In contrast, ViewSynth learns view synthesis in conjunction with learning keypoint-descriptor to improve keypoint matching performance.

\section{Methodology}
Here we describe our proposed depth image local feature learning framework \textbf{ViewSynth}.
We use a \textit{DaD} technique for learning keypoints and descriptors; while also learning 3D-structure-aware depth image representation using view synthesis.
\subsection{Learning keypoints and descriptors} \label{kplearningsubsection}
Given a depth image $I$, we first use VGG-16 \cite{simonyan2014vgg} up to the \texttt{conv\_4\_3} layer as a deep feature extractor $\Phi$ to extract features $F = \Phi(I)$.
$F \in \mathbb{R}^{h \times w \times f}$ is 8 times downsampled compared to $I$.
Here $h$, $w$ and $f$ refer to the height, width and channels of $F$ respectively.
$F_{i, j} \in \mathbb{R}^f$ indicates the features along location $(i, j)$.
Keypoints and their descriptors are extracted from $F$ using the joint keypoint detection-description technique described by \cite{dusmanu2019d2}.
Applying $L_2$ normalization to $F$ produces the descriptor at each spatial position, $D_{i, j} = F_{i, j} / ||F_{i, j}||_2$.
Potential keypoints are detected from $D$ with their respective soft detection scores $S \in \mathbb{R}^{h \times w}$ 
\cite{dusmanu2019d2} during training.
$S_{i, j}$ indicates how confident the local feature extraction method is in being able to correctly match the keypoint at $(i, j)$ in other images.
The hard feature detection technique \cite{dusmanu2019d2} is used during inference.

We learn local features by training the network $\Phi$ with pairs of depth images with some overlap, and by encouraging it to learn correct keypoint correspondences between depth images.
Given a pair of depth images $(I^{(1)}, I^{(2)})$ normalized to $[0, 1]$, their corresponding dense features $F^{(1)}$, $F^{(2)}$, soft keypoint scores $S^{(1)}, S^{(2)}$ and descriptors $D^{(1)}, D^{(2)}$ are first extracted as described earlier.
We define $\pi(F^{(j)})$ as the grid of spatial positions of $F^{(j)}$; and $\textbf{C}_{gt}$ as ground truth correspondences between the points visible in both images based on their 3D world coordinates.
For each correspondence $(c_1, c_2) \in \textbf{C}_{gt}$ such that $c_1 \in \pi(F^{(1)}), c_2 \in \pi(F^{(2)})$, we minimize the \textit{positive descriptor distance}, $p(c_1, c_2) = ||D^{(1)}_{c_1} - D^{(2)}_{c_2}||_2$ to encourage descriptor similarity between correct correspondences.
We also maximize \textit{negative descriptor distance}, the descriptor distance between the most confounding incorrect correspondences.
For $c_1$, we compute the descriptor distance of the most confounding incorrect correspondence in $I^{(2)}$: $n(c_1, c_2) = min_{k}{||D^{(1)}_{c_1} - D^{(2)}_k||}_2$, where $k \in \pi(F^{(2)})$, and $|| k - c_2 ||_2 > \tau$. 
$\tau$ defines a boundary around each correctly matched keypoint, within which we do not consider any point as a negative match.
Following \cite{dusmanu2019d2}, we use $\tau = 4$ pixels.
Similarly, we compute $n(c_2, c_1)$, the most confounding incorrect correspondence distance for $c_2$.

D2Net uses a triplet loss to encourage $p(c_1, c_2)$ to be smaller than $\min(n(c_1, c_2), n(c_2, c_1))$ upto some margin.
Interestingly, we observe that this loss often led to a model collapse \cite{wu2017sampling}, where all descriptors collapsed onto a singular representation in earlier phases of training.
We presume that the inherent difficulty associated with learning pose invariant representation from often noisy depth data, coupled with a high learning rate, and the hard negative sampling of D2Net led to this phenomenon.
We propose to use a contrastive loss \cite{yalecun2006con} to avoid this problem.
Unlike the triplet loss, the contrastive loss encourages the network to learn the exact same descriptor for a keypoint across depth images from any viewpoint.
This is desirable in ViewSynth, since we want the densely extracted features to encode information in a viewpoint invariant fashion.
The contrastive loss for $D^{(1)}_{c_1}$ is: $L_c(c_1, c_2) = 0.5 p(c_1, c_2)^2 + 0.5 \max(0, m - n(c_1, c_2))^2$.
Margin $m$ is empirically set to $1.5$ for all of our experiments.
Similarly, for $D^{(2)}_{c_2}$ we compute $L_c(c_2, c_1)$.
Finally, our proposed \dadlossfullname, $L_{\dadloss}$ is defined as:
\begin{equation}
    L_{\dadloss} = \frac{\sum_{c_1,c_2 \in \textbf{C}_{gt}} S^{(1)}_{c_1} S^{(2)}_{c_2} (L_c(c_1, c_2)+L_c(c_2, c_1))}{\sum_{c_1,c_2 \in \textbf{C}_{gt}} S^{(1)}_{c_1} S^{(2)}_{c_2}} .
\end{equation}
$L_{\dadloss}$ is a weighted average of the contrastive loss terms based on the keypoint scores.
Minimizing $L_{\dadloss}$ drives the relative scores of the correspondences with lower contrastive loss to get larger, and vice versa.
Simultaneously, $L_{\dadloss}$ also drives all contrastive losses to be smaller.
Optimization of $L_{\dadloss}$ is done wrt the parameters of feature extraction network $\Phi$.

Although Georgakis \etal \cite{georgakis2018end} use contrastive loss for learning descriptors - they learn keypoints and descriptors separately, while our method learns them jointly. Another key difference is: for each keypoint, \cite{georgakis2018end} only considers a single match on the other image, and penalizes depending on whether the match was correct or not. In contrast, our method considers \textit{all} possible matches on the other image, and decides the correct match and the most confounding incorrect match, which better lets our method have dissimilar descriptors for different keypoints.

\begin{figure*}[t!]
\centering
\includegraphics[width=1.0\textwidth]{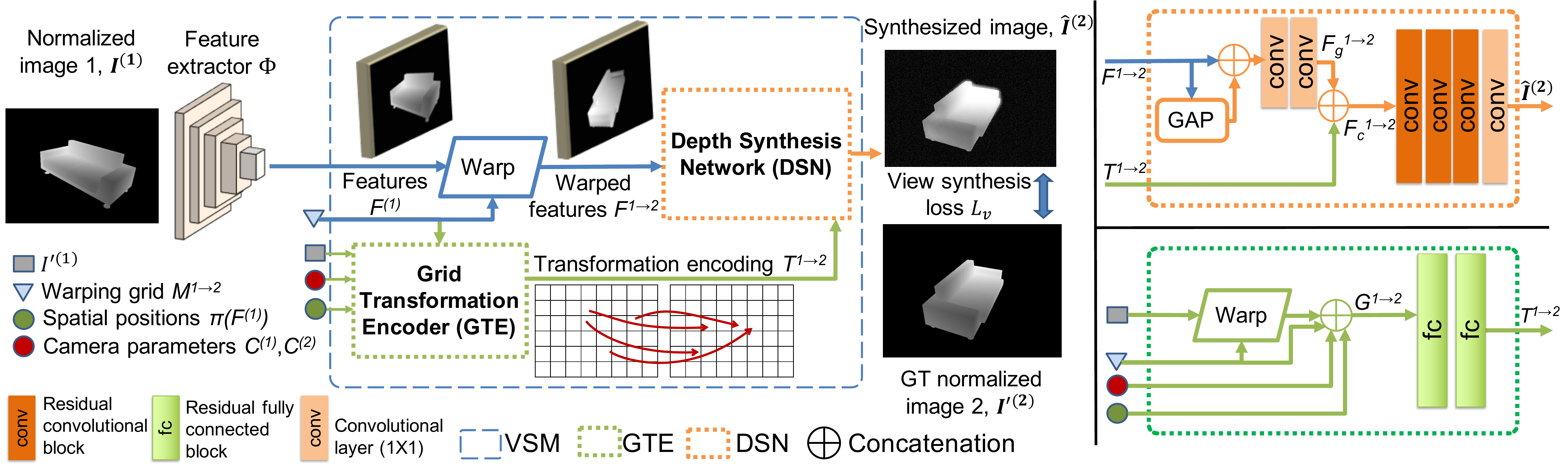}
\vspace{-0.3cm}
\caption{%
VSM takes in the dense representation $F^{(1)}$ of the depth image $I^{(1)}$ and the parameters related to pixel-wise transformation from $I^{(1)}$ to $I^{(2)}$, and synthesizes the normalized representation $\hat{I}^{(2)}$ from the view of $I^{(2)}$. See Sec. \ref{vsnsubsection} for details.
}
\vspace{-0.3cm}
\label{fig:vsnmodelfigure}
\end{figure*}
\vspace{-0.3cm}
\subsection{Learning from view synthesis}
\label{vsnsubsection}

Inspired by \cite{sitzmann2019scene,Suwajanakorn2018}, we hypothesize that learning 3D-structure-aware depth image representation using view synthesis can assist in learning local features more suitable for correct matching.
Intuitively, learning to synthesize views from unseen viewpoints can assist the keypoint detection and description process by encoding 3D structure information of the scene.
Consider depth images $I^{(1)}$ and $I'^{(2)}$ in Figure \ref{fig:vsnmodelfigure}. $I'^{(j)}$ is the 8 times downsampled version to $I^{(j)}$.
Some surfaces of the sofa observed in $I^{(1)}$ are occluded in $I'^{(2)}$.
We propose the \vsnfull~(\vsn) (Figure \ref{fig:vsnmodelfigure}) to learn to synthesize $I'^{(2)}$, from $F^{(1)}$, and $T^{1 \rightarrow 2}$: the relative transformations from $I^{(1)}$ to $I^{(2)}$.
We hypothesize that learning view synthesis will encourage $F^{(1)}$ to encode information beyond the visible pixels of the depth image, which can improve keypoint matching accuracy.
We summarize a high-level overview of VSM -
\begin{itemize}
    \item VSM first warps the features of $I^{(1)}$ onto $I^{(2)}$’s perspective.
    \item GTE encodes the $I^{(1)}$ to $I^{(2)}$ pixel-level transformation-related parameters along each spatial position.
    \item DSN then uses the warped features of $I^{(1)}$ and the transformation features from GTE to predict the estimated depth intensity along each pixel of $I^{(2)}$.
    \item ViewSynth is end-to-end differentiable, and learning view synthesis encourages the feature extractor $\Phi$ to learn to encode information about the visible and occluded parts of $I^{(1)}$.
\end{itemize}

During training, we use depth image pairs $(I^{(1)}, I^{(2)})$, with camera parameters $C^{(1)}$, $C^{(2)}$ respectively to learn view synthesis.
$C^{(j)}$ refers to the intrinsic and extrinsic parameters of camera $j$.
We compute the mapping grid $M^{1 \rightarrow 2}$, which indicates where each spatial location of $F^{(2)}$ is located in $F^{(1)}$.
$M^{1 \rightarrow 2}$ is mathematically defined on $C^{(1)}$, $C^{(2)}$ and computed using $I'^{(2)}$.
\vsn~uses $I'^{(2)}$ only to compute the mapping grid $M^{1 \rightarrow 2}$, and not directly as an input to any part of the neural network of \vsn.
Instead, \vsn~takes $F^{(1)}$, $\pi(F^{(1)})$, $I'^{(1)}$, $C^{(1)}$, $C^{(2)}$ as input, and utilizes $M^{1 \rightarrow 2}$ to synthesize the normalized depth image from $C^{(2)}$'s pose: $\hat{I}^{(2)}$.

\vsn~first uses the mapping grid $M^{1 \rightarrow 2}$ to warp the dense feature representation $F^{(1)}$ onto the image space of $F^{(2)}$ to obtain the warped representation, $F^{1 \rightarrow 2} = $ Warp$(F^{(1)}; M^{1 \rightarrow 2})$.
The \gtefull~(\gte) then computes $T^{1 \rightarrow 2}$.
Finally, the \dsnfull~(\dsn) uses $F^{1 \rightarrow 2}$ and $T^{1 \rightarrow 2}$ to synthesize $\hat{I}^{(2)}$.
\vsn~is needed only during training to learn view synthesis, and not required for keypoint-descriptor generation during inference.

\textbf{Grid Transformation Encoder (\gte):}
$L_{\dadloss}$ seeks to learn a viewpoint invariant depth image representation.
Hence, to synthesize $\hat{I}^{(2)}$, it is essential to use $T^{1 \rightarrow 2}$ to incorporate viewpoint specific information.
\gte~is designed to encode $T^{1 \rightarrow 2}$.
First, we compose a grid of transformation related parameters $G^{1 \rightarrow 2} \in \mathbb{R}^{h \times w \times f'_t}$.
Along spatial position $(i, j)$, $G^{1 \rightarrow 2}_{i, j}$ is created by concatenating: 1) Warp$(I'^{(1)}; M^{1 \rightarrow 2})(i, j)$, 2) $\pi({F^{(1)}})(i, j)$, 3) $M^{1 \rightarrow 2}_{i, j}$ and 4) the relative pose between $C^{(1)}$ and $C^{(2)}$.
Then, similar to \cite{zhou2016view} we encode $G^{1 \rightarrow 2}$ using two fully-connected residual blocks, and obtain $T^{1 \rightarrow 2} \in \mathbb{R}^{h \times w \times f_t}$.
We empirically set $f_t = 96$.

\textbf{Depthmap Synthesis Network (\dsn):}
\dsn~(Figure \ref{fig:vsnmodelfigure}) takes as input $F^{1 \rightarrow 2}$ and $T^{1 \rightarrow 2}$; and synthesizes $\hat{I}^{(2)}$.
First, we apply global average pooling (GAP) \cite{lin2013network} on $F^{1 \rightarrow 2}$.
GAP features are then concatenated across every spatial position of $F^{1 \rightarrow 2}$ and passed through two $ 1 \times 1 $ convolutions to obtain $F_g^{1 \rightarrow 2}$.
$F_g^{1 \rightarrow 2}$ captures spatial position specific local information, and also the global context captured by GAP.
Then we concatenate $F_g^{1 \rightarrow 2}$ with $T^{1 \rightarrow 2}$ along the channel dimension to obtain $F_c^{1 \rightarrow 2}$.
Finally, three residual convolutional blocks and a $1 \times 1$ convolutional layer follow to synthesize $\hat{I}^{(2)}$.
We use convolutional blocks to reason about the spatial neighborhoods of each pixel in the final phase of \dsn.

\textbf{View Synthesis Loss ($L_v$):}
We train \vsn~using the \vslfull~$L_v$, which encourages the similarity between $\hat{I}^{(2)}$ and $I'^{(2)}$.
We define $L_v$ as:
\begin{equation}
    L_v = \left. \left(\sum_{i \in P^{1 \rightarrow 2}}{|| \hat{I}^{(2)}_i -  I'^{(2)}_i ||_1} \right) \middle/ |P^{1 \rightarrow 2}| \right. .
\end{equation} Here,
$P^{1 \rightarrow 2}$ is the set of pixels in $I'^{(2)}$ that correspond to the 3D points contained within the camera view frustum of $I^{(1)}$, but possibly are occluded in $I^{(1)}$.
The overall loss $L$ for the ViewSynth is $L = L_{\dadloss} + \alpha L_v$, where we empirically set $\alpha=10$ throughout all experiments.

\section{Experimental evaluation} \label{experimental_eval_section}
\newcommand{\kprepo}{$\mathcal{R}$}
We evaluate the target tasks on three datasets: RGB-D dataset 7-Scenes (denoted as MSR-7) \cite{msr7_2013}, TUM RGBD-SLAM (3D object reconstruction subset) \cite{sturm12iros}, and CoRBS \cite{wasenmuller2016corbs}.
Summarized in Table \ref{tabledataset}, each dataset is a compilation of tracked sequences of real RGB-D camera frames of naturally occurring indoor scenes.

\textbf{Experimental Protocol}
We follow the same experimental setup as \cite{georgakis2018end} to evaluate local feature learning from depth images on the 3D correspondence matching and camera localization tasks.
Pairs of depth images captured 10 or 30 frames apart are used during training.
For evaluation, a \textit{reference 3D keypoint-descriptor repository}, \kprepo~is created from the training-set images.
This is done by extracting the 50 highest scoring keypoints and their descriptors from each image, and putting these keypoints' 3D world coordinates and their descriptors into \kprepo.
Next, we extract the 50 highest scoring keypoints from each test-set image and match them against the keypoints in \kprepo~based on the closest $L_2$ descriptor distance.
A match is considered correct if the 3D world coordinates of the matched keypoints are within a 3D distance threshold ($0.1m$, $0.25m$ or $0.5m$).
To evaluate camera localization, we use an experimental protocol similar to \cite{sattler2018benchmarking}.
Same as before, we match the keypoints of each test-set depth image to \kprepo, and then estimate the camera pose using the RANSAC based EPnP solver \cite{lepetit2009epnp,opencv_library}.
The camera localization accuracy is measured in different \textit{position error} and \textit{orientation error} thresholds. We evaluate on (0.5m, 2\degree), (1m, 5\degree) and (5m, 10\degree) thresholds.

\textbf{Baseline}
We use the SOTA depth local feature extractor Georgakis' method \cite{georgakis2018end}, and SOTA RGB local feature extractors R2D2 \cite{revaud2019r2d2} and D2Net \cite{dusmanu2019d2} adapted for depth modality as our baselines.
As the original D2Net led to the model collapse \cite{wu2017sampling} in all experimental setups, we add a modified D2Net baseline (mD2Net) which uses all negative sampling instead of the hardest negative sampling for descriptor learning.
For keypoint-descriptor extraction, we used the 3-scale detection setting \cite{dusmanu2019d2} for all D2Net baselines and ViewSynth.
We also add an additional R2D2 baseline \rdalgoms, which uses 3 scales (instead of 5) for keypoint aggregation similar to ViewSynth and the D2Net baselines.
While training the R2D2 baselines, the image pairs were resized to $256 \times 192$ to fit them into the GPU memory.
During evaluation, full resolution images were used for keypoint-descriptor extraction.
Finally, we add another baseline D2Net$_{L_{\dadloss}}$ which uses the D2Net architecture and $L_{\dadloss}$.
\begin{table}
\centering
\resizebox{0.8\columnwidth}{!}{
    \begin{tabular}{@{}c@{\hspace{.5em}}c@{\hspace{.5em}}c@{\hspace{.5em}}c@{\hspace{.5em}}c}
     \toprule
     Dataset $\backslash$ Property &\# of Scenes & \# of Sequences & Sensor Type & \# Training/Testing Images \\
     \midrule
    MSR-7 \cite{msr7_2013} & 7 & 18 & Kinect & 26K/17K \\
    TUM \cite{sturm12iros} & 11 & 55 & Kinect & 18K/4K\\
    CoRBS \cite{wasenmuller2016corbs} & 4 & 20 & Kinect v2 & 26K/6K \\
     \bottomrule
\end{tabular}
}
\vspace{.5em}
\caption{The datasets and their properties used in our experiments.} 
\label{tabledataset}
\end{table}

\begin{table}%
\centering
\resizebox{\columnwidth}{!}{
\begin{tabular}
{@{}cc@{\hspace{0.5em}}c@{\hspace{0.5em}}c@{\hspace{0.5em}}cc@{\hspace{0.5em}}c@{\hspace{0.5em}}c@{\hspace{0.5em}}cc@{\hspace{0.5em}}c@{\hspace{0.5em}}c@{\hspace{0.5em}}c@{}}
\toprule
MMA Threshold & \multicolumn{2}{c}{0.1m} & \multicolumn{2}{c}{0.25m} & \multicolumn{2}{c}{0.1m} & \multicolumn{2}{c}{0.25m} &\multicolumn{2}{c}{0.1m} & \multicolumn{2}{c}{0.25m}  \\ 
\midrule
\# of Frames Apart & 10 & 30 & 10 & 30 & 10 & 30 & 10 & 30 & 10 & 30 & 10 & 30 \\ \midrule
Dataset &\multicolumn{4}{c}{TUM} & \multicolumn{4}{c}{CoRBS} & \multicolumn{4}{c}{MSR-7} \\ \midrule[0.75pt]
D2Net \cite{dusmanu2019d2} & \multicolumn{4}{c}{Collapsed} & \multicolumn{4}{c}{Collapsed} & \multicolumn{4}{c}{Collapsed}  \\
mD2Net & 8.72 & 3.62 & 20.48 & 12.60 & 17.10 & 13.93 & 29.83 & 28.13 & 45.69 & 45.02 & 61.31 & 59.55\\
R2D2 \cite{revaud2019r2d2} & 20.84 & - & 37.34 & - & 42.08 & - & 51.26 & - & 61.55 & - & 66.30 & - \\
\rdalgoms~\cite{revaud2019r2d2}  & 16.74 & - & 33.24 & - & 34.70 & - & 43.15 & - & 50.07 & - & 55.60 & - \\
D2Net$_{L_{\dadloss}}$ & 33.38 & 23.93 & 53.19 & 45.82 & 56.73 & 51.53 & 71.24 & 66.65 & 79.87 & 80.35 & \textbf{89.84} & 90.30 \\
ViewSynth (ours) & \textbf{34.75} & \textbf{35.63} & \textbf{59.45} & \textbf{57.39} & \textbf{67.30} & \textbf{52.69} & \textbf{72.43} & \textbf{69.25} & \textbf{80.10} & \textbf{80.56} & 89.70 & \textbf{90.72} \\
\bottomrule
\end{tabular}%
}
\vspace{-0.7em}
\caption{Comparison of MMA on TUM, CoRBS, and MSR-7 datasets, trained on 10/30-frames-apart setting. Acronyms: mD2Net: modified D2Net; D2Net$_{L_{\dadloss}}$: D2Net with loss $L_{\dadloss}$; ViewSynth: D2Net$_{L_{\dadloss}}$+ $L_v$, our proposed method.
}
\label{tab:mmatable}
\end{table}

\begin{figure*}[t!]
\centering
\includegraphics[width=1.0\textwidth]{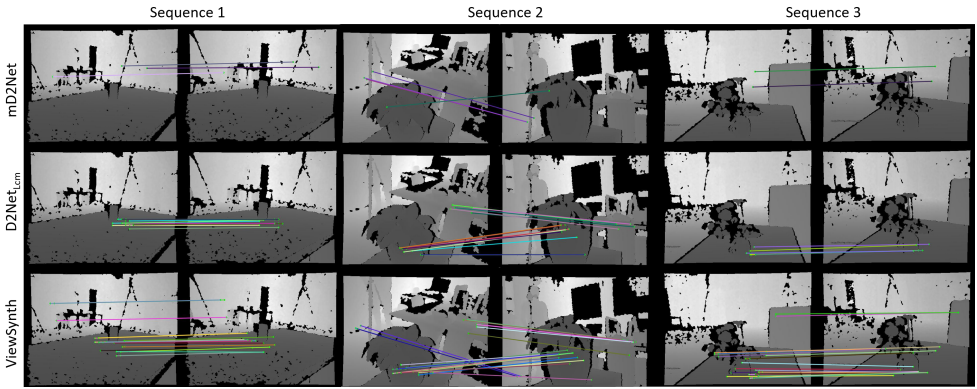}
\vspace{-0.3cm}
\caption{ViewSynth shows that learning view synthesis allows more correct keypoints matches between image pairs.}
\vspace{-0.3cm}
\label{fig:qualitativematching}
\end{figure*}

\begin{figure}
\centering
\includegraphics[width=0.95\linewidth]{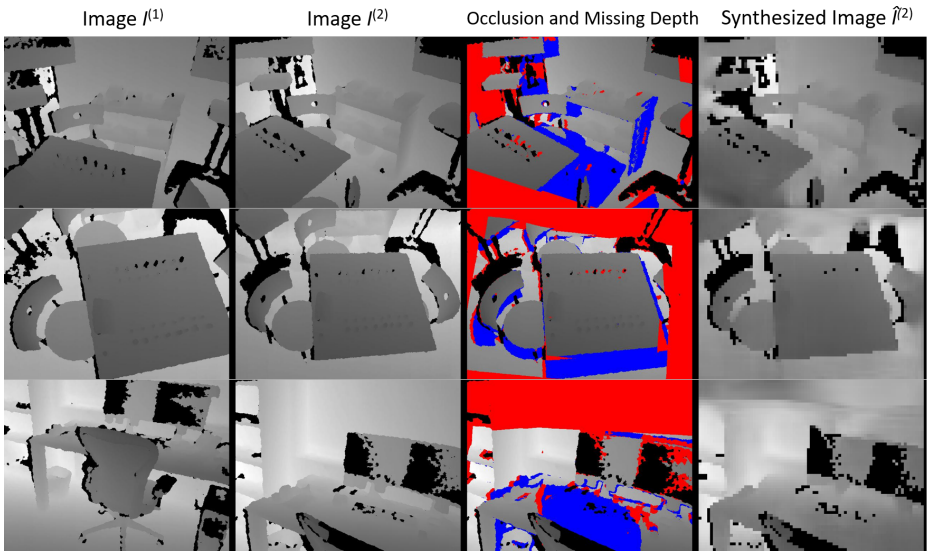}
\vspace{-0.3cm}
\caption{
View synthesis outputs on MSR-7 scenes. \textbf{\textcolor{blue}{Blue}} highlighted area indicates the parts of $I^{(2)}$ that are occluded in $I^{(1)}$. \textbf{\textcolor{red}{Red}} highlight indicates the change in pose between $I^{(1)}$, $I^{(2)}$ and the missing information in $I^{(1)}$. $\hat{I}^{(2)}$ shows that \vsn~can synthesize the depth views in the \textcolor{blue}{blue} occluded regions.
} 
\vspace{1mm}
\label{fig:viewsynthesis_supervision_viz}
\end{figure}

\textbf{Results}
Figure \ref{fig:qualitativematching} shows the qualitative results of keypoint matching, where ViewSynth obtains higher number of correct matching pairs than mD2Net and D2Net$_{L_{\dadloss}}$.
Table \ref{tab:mmatable} shows the mean matching accuracy (MMA) for each method in the 3D keypoint matching task, for $0.1m$, $0.25m$ and $0.5m$ 3D distance thresholds.
D2Net could not be trained as it faced the model collapse \cite{wu2017sampling} in all experiments.
ViewSynth outperforms all the listed baselines in most settings.
R2D2 had a convergence issue for 30-frames-apart training setting, for which it could not produce any keypoint.
Possibly, R2D2's sensitive approximated average precision failed to produce meaningful gradients for noisy depth images with large viewpoint variation.
ViewSynth beats Georgakis' method \cite{georgakis2018end} by 80.65 vs. 41.20 MMA in the MSR-7 dataset for 10-frames-apart training setting, which is the only experimental setting conducted by \cite{georgakis2018end} on this dataset.
ViewSynth outperforms the listed baselines in most settings of the camera localization task (Table \ref{tab:cameralocalizationtumcorbs}, \ref{tab:msr7cameralocalization}).
Figure \ref{fig:viewsynthesis_supervision_viz} demonstrates that ViewSynth is able to synthesize occluded surfaces, which it was trained to learn.

\begin{table}[t]
\centering
\resizebox{\columnwidth}{!}{%
\begin{tabular}
{@{}cccccccccccccc@{}}
\toprule
Threshold & \multicolumn{2}{c}{0.5m, 2°} & \multicolumn{2}{c}{1m, 5°} & \multicolumn{2}{c}{5m, 10°} & & \multicolumn{2}{c}{0.5m, 2°} & \multicolumn{2}{c}{1m, 5°} & \multicolumn{2}{c}{5m, 10°} \\ 
\midrule
\# of Frames Apart & 10 & 30 & 10 & 30 & 10 & 30 & & 10 & 30 & 10 & 30 & 10 & 30\\ 
\midrule
Method $\backslash$ Dataset&\multicolumn{6}{c}{TUM} & & \multicolumn{6}{c}{CoRBS} \\ 
\midrule[0.75pt]
D2Net \cite{dusmanu2019d2} & \multicolumn{2}{c}{Collapsed} & \multicolumn{2}{c}{Collapsed} & \multicolumn{2}{c}{Collapsed} & & \multicolumn{2}{c}{Collapsed} & \multicolumn{2}{c}{Collapsed} & \multicolumn{2}{c}{Collapsed} \\
mD2Net & 1.18 & 1.77 & 4.81 & 6.51 & 9.67 & 12.49 & & 1.90 & 4.18 & 6.85 & 11.56 & 13.40 & 18.51 \\
R2D2 \cite{revaud2019r2d2} & 7.44 & - & 18.70 & - & 26.40 & - & & \textbf{26.03} & - & \textbf{41.91} & - & \textbf{48.72} & - \\
\rdalgoms \cite{revaud2019r2d2} & 6.76 & - & 17.10 & - & 23.68 & - & & 25.28 & - & 38.93 & - & 44.48 & - \\
ViewSynth (ours) & \textbf{7.70} & \textbf{7.58} & \textbf{23.02} & \textbf{16.60} & \textbf{35.49} & \textbf{27.18} & & 8.19 & \textbf{8.57} & 23.36 & \textbf{30.29} & 47.78 & \textbf{50.52}\\ 
\bottomrule
\end{tabular}%
}
\vspace{-0.2cm}
\caption{Camera localization accuracy (\%) on TUM and CoRBS datasets, with 10/30-frames-apart training setting. For most localization correctness thresholds, our proposed method outperforms the SOTA.}
\vspace{-0.3cm}
\label{tab:cameralocalizationtumcorbs}
\end{table}

\textbf{Ablation}
We study the effect of $L_{\dadloss}$ and $L_v$ on both tasks.
Figure \ref{fig:qualitativematching} and Table \ref{tab:mmatable}, \ref{tab:msr7cameralocalization} show the efficacy of $L_{\dadloss}$, where D2Net$_{L_{\dadloss}}$ outperforms mD2Net in every case, while original D2Net fails to learn.
ViewSynth in addition uses \vsn~and $L_v$ to learn view synthesis, and outperforms D2Net$_{L_{\dadloss}}$ in most settings in Table \ref{tab:mmatable}, beating all the listed baselines.
\vsn~is more effective in the 30-frames-apart training setting compared to the 10-frames-apart training setting; as it learns view synthesis from larger viewpoint variation with more occlusion (Table \ref{tab:mmatable}).
This supports that - learning view synthesis encodes information beneficial for improving keypoint matching accuracy.
We also see in the camera localization task (Table \ref{tab:msr7cameralocalization}), that ViewSynth beats other methods in most settings.
Table \ref{tab:generalization} demonstrates the generalizability of ViewSynth, where it beats D2Net$_{L_{\dadloss}}$ in the keypoint matching task across different scenes of the same dataset, and across different datasets using the same or different depth sensors.
\begin{table}[t]
\centering
\resizebox{0.53\columnwidth}{!}{%
\begin{tabular}{@{}cc@{\hspace{0.5em}}cc@{\hspace{0.5em}}cc@{\hspace{0.5em}}c@{}}
\toprule
Threshold & \multicolumn{2}{c}{0.5m, 2°} & \multicolumn{2}{c}{1m, 5°} & \multicolumn{2}{c}{5m, 10°} \\ \midrule
\# of Frames Apart & 10 & 30 & 10 & 30 & 10 & 30 \\ \midrule
Method $\backslash$ Dataset &\multicolumn{6}{c}{MSR-7} \\ \midrule[0.75pt]
D2Net \cite{dusmanu2019d2} & \multicolumn{2}{c}{Collapsed} & \multicolumn{2}{c}{Collapsed} & \multicolumn{2}{c}{Collapsed} \\
mD2Net & 15.46 & 14.11 & 37.68 & 34.66 & 53.98 & 50.24 \\
R2D2 \cite{revaud2019r2d2} & \textbf{47.20} & - & 68.13 & - & 74.49 & - \\
\rdalgoms \cite{revaud2019r2d2} & 43.94 & - & 61.61 & - & 67.78 & - \\
D2Net$_{L_{\dadloss}}$ & 31.52 & 21.92 & 66.33 & \textbf{58.25} & 85.24 & \textbf{82.61} \\
ViewSynth (ours) & 34.60 & \textbf{23.83} & \textbf{70.09} & 57.04 & \textbf{86.67} & 80.34 \\ \bottomrule
\end{tabular}%
} 
\vspace{0.5 em}
\caption{Camera localization accuracy (\%) on MSR-7 dataset, with 10/30-frames-apart training setting. 
ViewSynth outperforms baselines for most localization correctness thresholds.
}
\vspace{-0.5cm}
\label{tab:msr7cameralocalization}
\end{table}
\begin{table}[t]
\centering
\resizebox{0.95\columnwidth}{!}{%
\begin{tabular}{ccc@{\hspace{0.1em}}cc@{\hspace{0.1em}}cc@{\hspace{0.1em}}c}
\toprule
\multicolumn{2}{c}{MMA Threshold} & \multicolumn{2}{c}{ $0.1m$ } & \multicolumn{2}{c}{ $0.25m$ } & \multicolumn{2}{c}{ $0.5m$ } \\
\midrule
Trained on & Tested on & D2Net$_{L_{\dadloss}}$ & ViewSynth & D2Net$_{L_{\dadloss}}$ & ViewSynth & D2Net$_{L_{\dadloss}}$ & ViewSynth \\
\midrule[0.75pt]
MSR-7 (w/o fire scene) & MSR-7 fire scene & 83.32 & \textbf{84.85} & 91.36 & \textbf{92.94} & 93.03 & \textbf{94.80} \\
MSR-7 & TUM & 27.93 & \textbf{29.05} & 51.89 & \textbf{53.88} & 67.36 & \textbf{69.08} \\
MSR-7 & CoRBS & 44.96 & \textbf{46.47} & 61.65 & \textbf{63.32} & 73.62 & \textbf{75.00} \\
\bottomrule
\end{tabular}
}
\vspace{0.5 em}
\caption{Generalizability of ViewSynth framework on the 3D keypoint matching task using MMA metric. ViewSynth generalizes better than D2Net$_{L_{\dadloss}}$ - to a new scene (row 1), new dataset with the same sensor (row 2), and across a dataset with different depth sensor (row 3) (see Table \ref{tabledataset} for the dataset details). All experiments are in the 30-frames-apart setting.
}
\label{tab:generalization}
\end{table}

\textbf{Discussion}
The quantitative (Table \ref{tab:mmatable}, \ref{tab:cameralocalizationtumcorbs}, \ref{tab:msr7cameralocalization}, \ref{tab:generalization}) and qualitative (Figure \ref{fig:qualitativematching}) results of the keypoint matching and camera localization tasks on different datasets show that ViewSynth outperforms the SOTA methods in most experimental settings.
R2D2 and \rdalgoms~perform reasonably in some settings, but they fail to learn when training image pairs contain large viewpoint variations.
In all cases, the original D2Net was not trainable due to the model collapse \cite{wu2017sampling}.
While mD2Net circumvents this using the all negative sampling for learning descriptors, it still leads to a poor performance.
D2Net$_{L_{\dadloss}}$ demonstrates the efficacy of the proposed loss $L_{\dadloss}$ by beating mD2Net in all cases.
ViewSynth in addition utilizes \vsn~and $L_v$ to learn view synthesis, and outperforms D2Net$_{L_{\dadloss}}$ and all the other baselines for both tasks in most settings.
These results assert the effectiveness of learning view synthesis for keypoint-descriptor extraction from depth images - towards the 3D keypoint matching and camera localization tasks.

\section{Conclusion}
\vspace{-1mm}
We show that the SOTA RGB keypoint detection-description methods either are not trainable (D2Net \cite{dusmanu2019d2}), or do not perform well (R2D2 \cite{revaud2019r2d2}) in the depth image modality.
Towards improving keypoint matching in the depth modality, we propose a framework \textit{ViewSynth} to learn view synthesis in conjunction with learning keypoint-descriptor from depth images in a joint fashion.
We propose the \dadlossfullname, $L_{\dadloss}$, to learn keypoints and descriptors jointly.
We show that, learning view synthesis of depth images from different viewpoints using our proposed \vsnfull~(\vsn) and the \vslfull, $L_v$, encourages the network to encode information that improves the performance in keypoint matching.
ViewSynth outperforms the SOTA in the 3D keypoint matching and camera localization task across the MSR-7, TUM and CoRBS datasets in most cases.
We also demonstrate the generalizability of ViewSynth in 3D keypoint matching across different datasets.
These evaluations attest the efficacy of ViewSynth in learning keypoints and descriptors for 3D keypoint matching and camera localization.

\textbf{Acknowledgement}
Part of the efforts from Jisan Mahmud and Jan-Michael Frahm is supported by NSF grant No. IIS-1816148.

\bibliography{bmvc_review}
\end{document}